\documentclass[lettersize,journal]{IEEEtran}
\usepackage{amsmath,amsfonts}
\usepackage{algorithmic}
\usepackage{algorithm}
\usepackage{array}
\usepackage[caption=false,font=normalsize,labelfont=sf,textfont=sf]{subfig}
\usepackage{textcomp}
\usepackage{stfloats}
\usepackage{url}
\usepackage{verbatim}
\usepackage{graphicx}
\usepackage{cite}
\usepackage{booktabs}
\usepackage{multirow}
\usepackage{pifont}       
\usepackage{bbding}       
\usepackage{fontawesome}  
\usepackage{caption}
\usepackage{graphicx}
\usepackage{float} 
\usepackage{subcaption}
\newcommand{\schemename}{IoTGen}

\begin{document}


\title{Synthetic User Behavior Sequence Generation with Large Language Models for Smart Homes}

\author{IEEE Publication Technology,~\IEEEmembership{Staff,~IEEE,}
\author{%
  Zhiyao Xu$^\dagger$, 
  Dan Zhao\thanks{$^{\mathsection}$Corresponding author: Dan Zhao (zhaod01@pcl.ac.cn), Yong Jiang(jiangy@sz.tsinghua.edu.cn)}$^{\ddagger\mathsection}$, 
  Qingsong Zou$^\dagger$,
  Jingyu Xiao$^\ast$,
  Yong Jiang$^{\dagger\ddagger\mathsection}$,
  Zhenhui Yuan$^\natural$,
  Qing Li$^\ddagger$\\
  $^\dagger$Tsinghua Shenzhen International Graduate School, China \\ $^\ddagger$PengCheng Laboratory, China\\
  $^\ast$The Chinese University of Hong Kong, China \\
  $^\natural$University of Warwick, United Kingdom\\
  \texttt{zhixu9557@gmail.com}; \texttt{zouqs21@mails.tsinghua.edu.cn};\\
  \texttt{\{liq,zhaod01\}@pcl.ac.cn}; \texttt{jiangy@sz.tsinghua.edu.cn}; \\
  \texttt{whalexiao99@gmail.com}; \texttt{zhenhui.yuan@warwick.ac.uk}
}

\thanks{This paper was produced by the IEEE Publication Technology Group. They are in Piscataway, NJ.}
\thanks{Manuscript received April 19, 2021; revised August 16, 2021.}}



\maketitle

\begin{abstract}
In recent years, as smart home systems have become more widespread, security concerns within these environments have become a growing threat. Currently, most smart home security solutions, such as anomaly detection and behavior prediction models, are trained using fixed datasets that are pre-collected. However, the process of dataset collection is time-consuming and lacks the flexibility needed to adapt to the constantly evolving smart home environment. Additionally, the collection of personal data raises significant privacy concerns for users. Lately, large language models (LLMs) have emerged as a powerful tool for a wide range of tasks across diverse application domains, thanks to their strong capabilities in natural language processing, reasoning, and problem-solving. In this paper, we propose an LLM-based synthetic dataset generation \schemename \ framework to enhance the generalization of downstream smart home intelligent models. By generating new synthetic datasets that reflect changes in the environment, smart home intelligent models can be retrained to overcome the limitations of fixed and outdated data, allowing them to better align with the dynamic nature of real-world home environments.Therefore Specifically, we first propose a Structure Pattern Perception Compression (SPPC) method tailored for IoT behavior data, which preserves the most informative content in the data while significantly reducing token consumption. Then, we propose a systematic approach to create prompts and implement data generation to automatically generate IoT synthetic data with normative and reasonable properties, assisting task models in adaptive training to improve generalization and real-world performance.
\end{abstract}


\section{Introduction}
\IEEEPARstart{T}{he} rapid growth of IoT solutions has driven an unprecedented rise in the number of smart devices within homes, with estimates suggesting that this number will reach approximately 5 billion by 2025 \cite{iot-analytics}. Smart home systems connect a wide array of IoT devices, enabling them to monitor users' living environments, capture their instructions and behavioral patterns, and interact directly with their living spaces. While these systems offer significant convenience, their close integration with users' private lives also presents substantial risks to both security and privacy \cite{anderson2014synthetic}. 
To enhance user convenience and safety in smart homes, these systems have integrated intelligent models to detect harmful or abnormal behaviors. They can also make recommendations or automatically take actions based on user behavior or contextual information, making this approach a common solution paradigm for home IoT \cite{xiao2024make, xiao2023know}.

However, most smart home intelligent models are trained in a one-off manner using pre-collected datasets, which require significant time and labor to gather. In the real world, user behavior can be influenced by factors such as season, lifestyle, and work status, leading to both subtle and significant changes over time. Although these pre-collected IoT datasets are derived from real-world scenarios, they capture only a small snapshot of a user's extended usage period. As a result, they lack the ability to account for dynamic changes and may lose their effectiveness over time, potentially causing significant issues when these models are applied in real-world settings. For example, behavior considered normal in the original context may be deemed abnormal or harmful in a different scenario, while behavior previously classified as abnormal may become normal in a new context. This mismatch can result in significantly higher rates of missed detections and false alarms. Another example of this limitation is the use of fixed patterns learned in a single, static scenario to predict user behavior in dynamic environments, which proves both inefficient and unreliable. Recent studies have highlighted these shortcomings. For instance, the author of the anomaly detection model ARGUS \cite{rieger2023argus} noted that the model struggles to adapt to significant changes, leading to an excessively high false positive rate and an inability to handle unforeseen events. Similarly, a study on using graph networks for behavior prediction \cite{xiao2023know} pointed out that such models lack adaptability to specific environments and face challenges in managing the complexities introduced by multiple users.

Conventional solutions to these challenges typically involve continuously collecting user data for model retraining. However, this approach has several drawbacks. First, gathering sufficient data often requires days or weeks, during which the outdated model remains in use, leading to degraded performance. Second, frequent tracking and data collection can raise significant privacy concerns among users. Third, the collected data often requires format conversion and pre-processing, which can introduce inefficiencies and disrupt system consistency.

To realize smart home intelligent models that are adaptable and flexible in real-world user environments, the following key questions must be addressed: 1) How to make the smart home system adapt to the ever-changing and unpredictable real home scenarios? 2) How to update the system without infringing user privacy and security? 3) How to ensure the reliability and timeliness of the smart home intelligent models?

Recently, the rapid advancement of large language models (LLMs) has offered a promising new approach to addressing these challenges. LLMs have demonstrated exceptional capabilities in knowledge retention, semantic understanding, and simulated reasoning. For instance, Interaction2Code \cite{xiao2024interaction2code} enables the generation of fully dynamic websites using LLMs, while ChatIoT \cite{gao2024chatiot} leverages LLMs to create TAP automation rules.

 In this paper, we propose an LLM-based IoT synthetic data generation (\schemename) framework to simulate new scene data, enhancing task model generalization for smart home systems and offering a solution for building open-world smart home systems. 
Building such a framework, however, faces the following challenges.

\begin{itemize}
\item{\textbf{Challenge 1}: Enabling efficient use of LLMs generation. LLMs face challenges such as maximum input length constraints, high inference costs, and prolonged inference time due to long inputs. Additionally, the lost-in-the-middle problem, where important information in the middle of long inputs is overlooked, further hinders their efficient use in smart homes.}
\item{\textbf{Challenge 2}: Creating a paradigm for LLMs to understand IoT smart home instruction and generate IoT synthetic data. This requires a well-designed set of LLM instructions to enable LLMs to generate high-quality synthetic data in a standardized format.}
\item{\textbf{Challenge 3}: Ensuring the adaptability of the synthetic data to the new scene and the consistency with the original data. The synthetic data should incorporate new pattern information relevant to the updated scene while maintaining the original user behavior habits and fundamental behavior patterns.}
\end{itemize}

In \schemename, we address the above challenges by the following designs.
First, in IoT scenarios, even sequences with high similarity can contain significantly different and contextually important information. To address this, we introduce the Structure Pattern Perception Compression (SPPC) method, which leverages autoencoders to assess the pattern information within data sequences and quantify their importance. This method effectively reduces token consumption while preserving the dataset's information integrity. Then, we propose a systematic approach for prompt construction, incorporating key elements such as role, task definition, requirement, scene information, and data information. By setting specific requirements and learning from original data, we can automatically generate IoT synthetic data with normative and reasonable properties.


\begin{figure*}[ht]
\centering
\includegraphics[width = .95\textwidth]{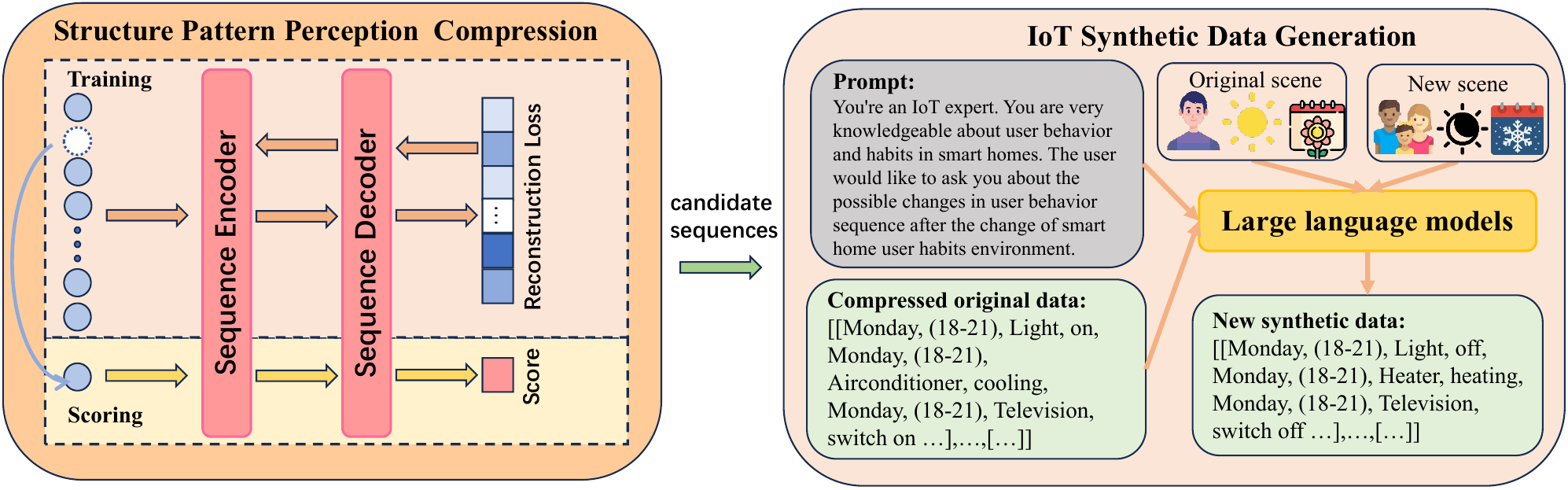}
\caption{The Overview of \schemename.}
\label{fig:flow}
\end{figure*}

\section{RELATED WORK}
This paper aims to generalize the task model in the smart home scenario through large language model data synthesis, which requires the use of data synthesis, large language model and other technologies, and mainly focuses on the two types of smart home models: anomaly detection and behavior prediction. 

\subsection{Data Synthesis for IoT}
Data synthesis is a widely explored topic in the field of machine learning and data science, enabling the generation of tabular data, images, text, audio, and more. Data synthesis is particularly essential in data-constrained domains, such as IoT scenarios, where real-world data is either unavailable or insufficient. For example, Jason W. Anderson et al. \cite{anderson2014synthetic} proposed using synthetic XML data to address challenges such as  possible user privacy violations and multi-platform conflicts. SA-IoTDG \cite{mondal2022situation} employed hidden Markov models to simulate real data and synthesize IoT traffic data with certain situational awareness capabilities. For sequence data generation, some recent studies \cite{redvzovic2017ip}, \cite{yin2022practical}, \cite{patki2016synthetic} have focused on generating packet-level data to represent sequences. For example, in \cite{redvzovic2017ip}, the authors used a hidden Markov model to implement an IP traffic generator. Yin et al. \cite{yin2022practical} used a time series model to generate time-spaced traffic blocks for each 5-tuple flow. IoTGemini \cite{li2024iotgemini} proposed using a sequential GAN to generate synthetic traffic with high fidelity. 
However, existing methods have significant limitations. On the one hand, they either narrowly focus on generating a single IoT file in a specific format or on producing synthetic traffic, leaving the synthesis of user behavior sequences largely unaddressed in the IoT domain. On the other hand, these methods are designed to replicate data within predefined scenarios, lacking the flexibility to generate data for previously unknown scenarios or to adapt dynamically to new conditions. Our approach addresses these gaps, providing a more comprehensive and adaptable solution for IoT data synthesis.

\subsection{Large Language Model}
The emergence of large language models (LLMs) has revolutionized natural language processing (NLP), achieving remarkable success across various tasks. Beyond NLP, LLMs have demonstrated immense potential in other fields, offering transformative capabilities.

Trained on vast amounts of data, LLMs possess exceptional semantic understanding, extensive knowledge reserves, and the ability to follow instructions effectively. They excel in open-ended tasks, often performing impressively even in zero-shot scenarios. For instance, Interaction2Code \cite{xiao2024interaction2code} showcases how an untrained LLM can generate fully functional dynamic interactive web pages based solely on instruction prompts. ChatIoT \cite{gao2024chatiot} highlights the application of LLMs in generating TAP automation rules for IoT systems.

A notable limitation of LLMs lies in their handling of excessively long token sequences. The limited token capacity restricts their ability to process very long inputs, and the inference speed increases significantly with token length due to the quadratic complexity of the attention mechanism. Moreover, excessively long token sequences can cause LLMs to lose focus on critical information, often leading to a ``lost-in-the-middle'' \cite{liu2024lost} effect where important details in the middle of the sequence are overlooked.


\subsection{Smart Home Models}
The increasing number and diversity of IoT devices has expanded the potential attack surface of many IoT systems, making them more vulnerable to security threats. Malicious attackers use various technical means to induce abnormal behavior or disrupt normal execution of IoT processes, potentially causing economic losses or even physical harm to users in smart homes. To mitigate these risks, various IoT anomaly detection systems have been proposed in recent years. For example,ARGUS \cite{rieger2023argus} employs an autoencoder composed of GRU to learn normal behavior and thereby identify contextual anomalies. SmartGuard \cite{xiao2024make} incorporates multi-dimensional time encoding and error-guided masking mechanisms to handle time-related sequence anomalies. 

User behavior prediction is another valuable research direction for enhancing both user safety and convenience in smart homes. By capturing and learning patterns from normal interaction data between users and IoT devices, it enables the prediction of potential user interaction needs, allowing smart home systems to provide proactive suggestions or assistance, thereby improving their intelligence and comfort. Furthermore, by accurately predicting normal behavior, any deviation from these patterns can be promptly flagged as potentially abnormal, helping to identify and mitigate potential risks, thereby significantly enhancing the safety of smart homes. For instance, IoTBeholder \cite{zou2023iotbeholder} uses the LSTM network with attention mechanism to predict user behavior. DeepUDI \cite{xiao2023user} and SmartUDI \cite{xiao2023know} use relational gated graph neural networks, capsule neural networks and contrastive learning to model users’ routines and intents for user behavior prediction.


In smart homes, contexts evolve dynamically over time due to factors such as seasonal changes, shifts in user habits and lifestyles, or alterations in daily routines and work conditions. Models trained on static, pre-collected datasets may struggle to adapt to these changes, leading to misinterpretations of device and user behaviors and inaccurate judgments about whether activities are normal or abnormal.

Existing anomaly detection and behavior prediction solutions struggle to effectively address these dynamic changes. They typically require frequent retraining to maintain accuracy, yet retraining necessitates collecting new datasets, which is often time-consuming and resource-intensive. Moreover, these models still face challenges in adapting to rapidly shifting scenarios and generalizing across diverse or unforeseen contexts. Their limited integration of contextual information further undermines reliability, increasing the likelihood of errors in managing the complexities of smart home systems. A promising solution to this challenge is synthesizing adaptive datasets that reflect evolving scenarios, enabling models to be retrained without the costly process of continuously collecting new real-world data.

\section{Overview}
\subsection{Problem Formulation}
Let $\mathcal{D}$ denote the set of IoT devices, $\mathcal{C}$ denote the set of device controls.

\noindent(\textbf{Behavior}) A behavior $b=(t, d, c)$ is a 3-tuple consisting of time stamp $t$, device $d \in \mathcal{D}$ and device control $c \in \mathcal{C}$. For example, behavior \textit{b = (2022-08-04 18:30, air conditioner, air conditioner:switch on)} describes the behavior ``\textit{swich on the air conditioner}'' at 18:30 on 2022-08-04.


\noindent(\textbf{Behavior Sequence}) A behavior sequence $s=[b_{1}, b_{2}, \cdots, b_{n}]$ is an ordered list of behaviors arranged by timestamps, where $n$ is the length of the sequence.  The dataset, denoted as $\mathcal{S}$, comprises a collection of such behavior sequences.



\subsection{Methodology}
We propose the \schemename \, a framework that uses a large language model to generate synthetic data for adaptive training of task models, in order to achieve a truly flexible and generalized smart home system. \schemename \ consists of two main modules: an Structure Pattern Perception Compression (SPPC) module, and an IoT Synthetic Data Generation module. The Structure Pattern Perception Compression module measures the richness of structural pattern information in IoT sequence data through an autoencoder and assigns importance scores to the sequences. By ranking these scores, the module filters and preserves truly representative IoT sequence data, thereby enhancing the efficiency of data generation. The IoT Synthetic Data Generation module takes the IoT sequence data from the Structure Pattern Perception Compression module as input, adds corresponding instructions and dictionaries, and generate IoT sequence data for new scenes. The generated data can then assist task models in adaptive training to improve the generalization and continuous reliability of the models.

\begin{table*}[t]
\small
\setlength{\tabcolsep}{0.1em}
\caption{\large System message design for \schemename.}
\label{tab:findings}
\begin{tabular*}{\linewidth}{@{}l|l@{}}
\toprule
Elements                          & Content                                                                                                                                                                                                                                                                                                                                                                                           \\ \midrule
\multirow{1}{*}{Role}  & \begin{tabular}[c]{@{}l@{}}You're an IoT expert. You are very knowledgeable about user behavior and habits in smart homes. The user would like to ask \\ you about the possible changes in user behavior sequence after the change of smart home user habits and environment.\end{tabular}  \\ \cmidrule(r){1-2}
\begin{tabular}[c]{@{}l@{}}Task \\ Definition \end{tabular}      &\begin{tabular}[c]{@{}l@{}}The user will provide you with the user's previous life environment and the changed environment, the user's previous behavior \\ sequence, and a set of devices and device states. And the user hope that you can use the devices and device states in the set \\ to generate possible user behavior sequences after the environment changes based on the original user behavior sequence.\end{tabular} \\\cmidrule(r){1-2}
        \multirow{12}{*}{Requirements} & \begin{tabular}[c]{@{}l@{}}Please strictly follow the correspondence between the devices and device states in the set to generate. Do not generate device \\ states that do not match the device. 
        \end{tabular} \\\cmidrule(r){2-2}
        & \begin{tabular}[c]{@{}l@{}}
         You can add some devices that users have not used before to better adapt to changes in the environment. 
         \end{tabular} \\\cmidrule(r){2-2}
        & \begin{tabular}[c]{@{}l@{}}
        Please modify or delete all unreasonable behaviors in the new environment.
        \end{tabular} \\\cmidrule(r){2-2}
        & \begin{tabular}[c]{@{}l@{}}
        Please consider as many new devices as possible in the new environment.
        \end{tabular} \\\cmidrule(r){2-2}
        & \begin{tabular}[c]{@{}l@{}}
        Please make sure that the generated sequence still contains 10 consecutive behaviors and there are forty elements in total.
        \end{tabular} \\\cmidrule(r){2-2}
        & \begin{tabular}[c]{@{}l@{}}
        Please ensure that the total number of generated behavior sequences is roughly equal to the total number of original behavior \\ sequences.
        \end{tabular} \\\cmidrule(r){2-2}
        & \begin{tabular}[c]{@{}l@{}}
        Please make modifications in the original sequence. The generated new behavior sequence set is also in the format \\ of [[...], [...], ...].
        \end{tabular} \\\cmidrule(r){1-2}
        
\multirow{2.5}{*}{\begin{tabular}[c]{@{}l@{}}Scene \\ Information \end{tabular}} & 
        \begin{tabular}[c]{@{}l@{}}The previous environment is {$E_1$}.
        \end{tabular}  \\ \cmidrule(r){2-2}
        & \begin{tabular}[c]{@{}l@{}}The changed environment is {$E_2$}. 
        \end{tabular}  \\ \cmidrule(r){1-2}

\multirow{2.5}{*}{\begin{tabular}[c]{@{}l@{}}Data \\ Information \end{tabular}} & 
        \begin{tabular}[c]{@{}l@{}}The user's previous sequence of behavior: \{user sequence\}.
        \end{tabular}  \\ \cmidrule(r){2-2}
        & \begin{tabular}[c]{@{}l@{}}The set of the possible device and device states: \{fr device control dict\}. 
        \end{tabular}  \\ 
        \bottomrule
\end{tabular*}
\end{table*}

\section{Structure Pattern Perception Compression}
In smart home scenarios, users frequently interact with IoT devices. For example, the SmartSense dataset contains 2,000 FR data entries and 8,000 US data entries. After converting these entries to text, the resulting data spans 914k to 3,657k tokens. Such extensive text contains way too much redundancy when used as prompt input. When combined with external knowledge, documents, or dictionaries, the prompt length often grows to an unmanageable extent. This poses two key challenges: first, excessively long prompts can significantly slow down the response speed of large language models (LLMs) and may even exceed their processing limits. Second, longer prompts lead to higher LLM call costs. 

In fact, in many cases, the dataset itself contains significant informational redundancy. To address the challenges outlined above, a Structure Pattern Perception Compression module is proposed to achieve prompt compression. Prompt compression aims to reduce the length of the input prompt by minimizing the number of tokens provided to the LLM without compromising the quality of its outputs. This approach enhances the response speed of the LLM while simultaneously reducing the computational cost associated with its usage.

Prompt compression offers several benefits. First, it improves inference speed by addressing the primary source of response delay in large language models, i.e., the prefill computation stage. Compressing the prompt significantly reduces this delay. Second, it lowers inference costs. The cost of large model inference services is closely tied to the number of input and output tokens, as both the prefill computation of prompts and the decoding of responses scale with token counts. Reducing the number of input tokens through compression effectively reduces these costs. Third, prompt compression significantly enhances the quality of generated outputs. Excessively lengthy prompts often lead to the ``lost in the middle'' problem, where critical contextual information gets overlooked as the model processes long inputs. This issue can severely degrade the quality of the model's responses. Additionally, irrelevant or noisy inputs, such as common noise data in IoT datasets, further distract the model and reduce its performance. By removing redundancy and mitigating noise, prompt compression allows the model to better focus on relevant details, effectively capture user habits, and generate more accurate and reliable results.


\begin{figure}[ht]
\centering
\includegraphics[width = .35\textwidth]{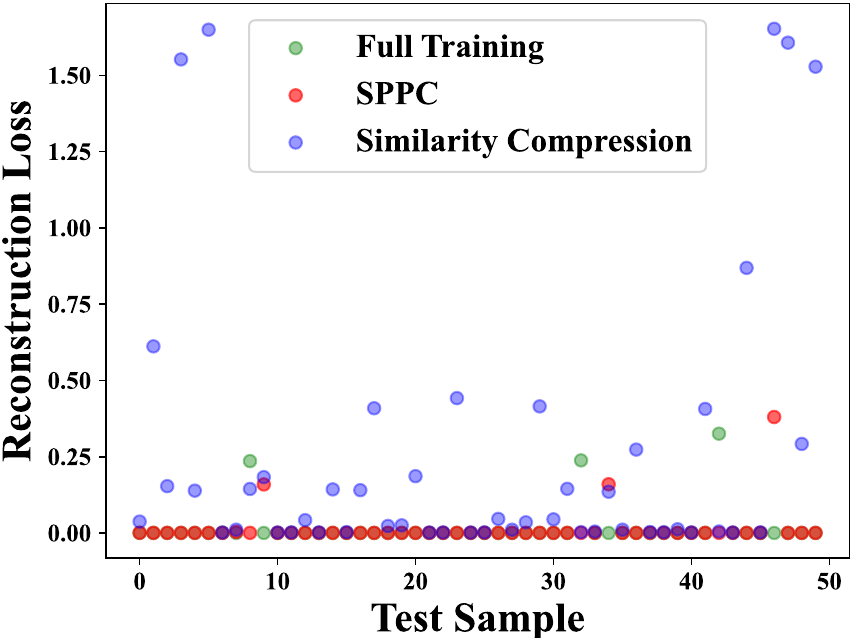}
\caption{Reconstruction Loss of Model Trained with Full/Compressed Data on The Test Dataset (TOP 50).}
\label{fig:rec_compare}
\end{figure}

However, using similarity index, which is a commonly used approach for compressing and merging similar sequence data, is inadequate in the context of smart home applications. The task requirements in this domain are highly specific, as the goal is to learn all possible user behavior patterns rather than merely approximate patterns. This is crucial for enabling accurate anomaly detection and behavior recommendation since capturing as many distinct patterns as possible is essential to define the boundaries of normal behavior comprehensively.

In smart home scenarios, behavior sequences that appear similar may represent distinct and equally informative patterns. For instance, a sequence of actions related to watching TV at night and another for watching TV during the day might differ only in their timestamps, resulting in very similar similarity indices. However, both sequences together can indicate that the user may watch TV throughout the day. Over-aggregation of such sequences—where one is deemed redundant and removed—can lead to significant deviations in the model's understanding of user behavior. 


We conduct an experiment to illustrates the drawback of similarity-based method. 
We designed a basic experiment that is common in anomaly detection tasks. The goal of this experiment is to train an anomaly detection model using normal samples and evaluate the learning effect of the model. Three different training sets were used in the experiment: Original dataset: The entire dataset was used as the training set; Compressed dataset (similarity method): The compressed dataset obtained by the similarity method was used as the training set; Compressed dataset (SPPC method): The compressed dataset obtained by the SPPC method we proposed was used as the training set. Then use one of the three training sets mentioned above to train the anomaly detection model. During the testing phase, the model is evaluated using the same test set, which only contains normal samples. The learning effect of the model is evaluated by calculating the reconstruction loss of the model for normal samples on the test set. A lower reconstruction loss indicates that the model has a better learning effect on normal samples.

As shown in Fig.~\ref{fig:rec_compare}, the model trained on the dataset compressed using a similarity-based method exhibits a significant increase in reconstruction loss. This result indicates that the similarity-based method is ineffective at compressing the dataset while adequately preserving its informative parts.  


\begin{figure}[ht]
\centering
\includegraphics[width = .35\textwidth]{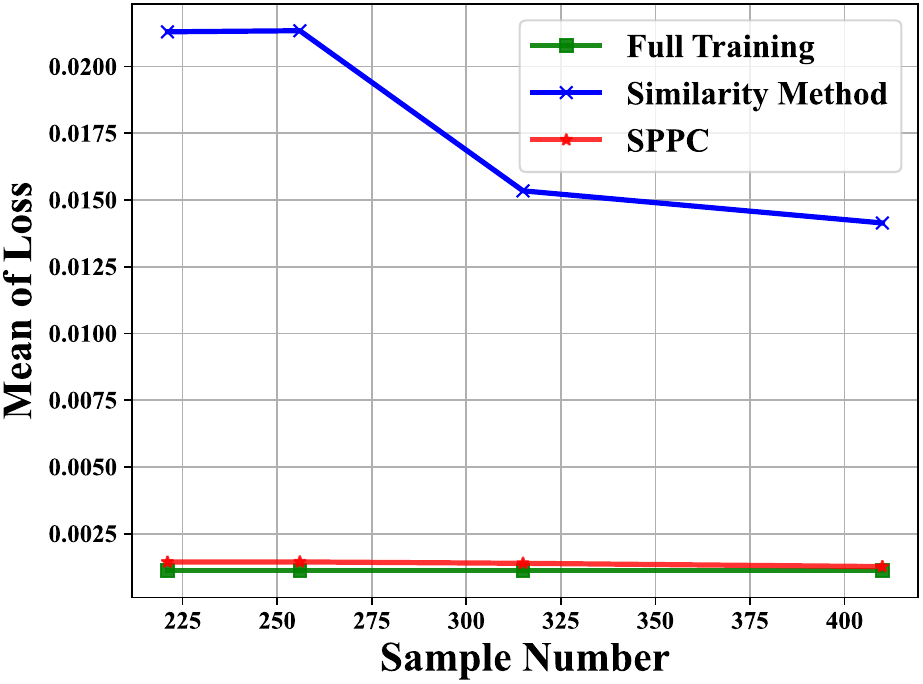}
\caption{Mean of Loss with Model Trained at Different Compression Levels on The Test Dataset.}
\label{fig:mean}
\end{figure}

\begin{figure}[ht]
\centering
\includegraphics[width = .35\textwidth]{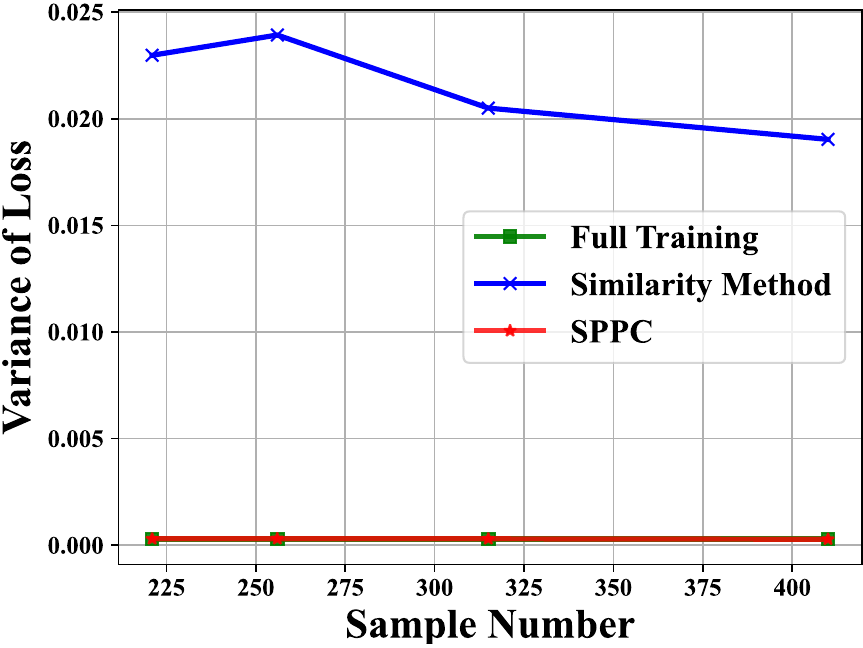}
\caption{Variance of Loss with Model Trained at Different Compression Levels on The Test Dataset.}
\label{fig:variance}
\end{figure}

In this paper, we propose a structure pattern perception compression (SPPC) method that can effectively compress IoT user behavior sequences. 

Given a behavior sequence dataset $S=\{s_1, s_2, ..., s_i, ...\}$, to measure the importance of a sequence $s_i\in S$, we first take $s_i$ out from $S$. Then, the remaining dataset $S \setminus \{s_i\}$ is fed to an autoencoder $T$ for seq2seq reconstruction training. Once the autoencoder $T$ is trained, we use it to reconstruct  $s_i$.  If $s_i$ is accurately reconstructed, this implies that its information is well-represented by other sequences, indicating lower significance. Conversely, if $s_i$ cannot be well reconstructed, this indicates that the information contained in $s_i$ is not redundant and should be preserved. 

Note that the reconstruction error of $s_i$ naturally serves as an importance score, which can be leveraged for sequence selection and compression in subsequent processing.

Figures \ref{fig:rec_compare}, \ref{fig:mean}, and \ref{fig:variance} present the reconstruction loss, mean loss, and loss variance for the full training method, the similarity-based method, and SPPC. As observed, SPPC achieves performance comparable to that of the full dataset training model across all three metrics while significantly outperforming the similarity-based method. Specifically, SPPC yields a reconstruction loss that is very close to that of the full dataset, whereas the similarity-based method exhibits a substantially higher loss. Moreover, SPPC achieves a mean loss and variance that are much lower than those of the similarity-based method and closely align with the full dataset results. These results demonstrate that SPPC more effectively preserves essential sequence information while achieving superior compression performance compared to the similarity-based approach.


\section{IoT Synthetic Data Generation}
Considering the remarkable performance of LLMs across various domains—particularly their vast knowledge base and strong semantic understanding—we opted to leverage them for IoT data synthesis. To achieve this, we design a structured and systematic workflow with tailored prompt instructions, comprising four key components: role, task definition, requirement, scene information and data information. The following is the detailed introduction.

First, to improve generation efficiency and quality, the original IoT data is first compressed by the SPPC module. Then, to facilitate the understanding and reasoning of the LLM, we transform the compressed sequence data from item to text description according to the smart home device dictionary, which records all devices that may appear in the smart home as well as device status and time information, and convert it into a text sequence, thus completing the data preparation stage.

Second, as shown in Figure \ref{fig:flow}, we need to proceed to the input stage. The input content includes: 1) the original IoT sequence data after data preparation, which is used to provide the original user behavior patterns and habits, as well as the basic IoT device interaction logic; 2) the smart home device dictionary, including timestamp sets, device sets and device interaction sets, which are used to provide the scope of generating new data and enhance the understanding of old data; 3) descriptions of new and old scenes, which are used to indicate the generation direction of the large model. For example, from single-person residence to multi-person residence, from daytime activities to nighttime activities, from spring to winter; 4) instructions, which are used to standardize and improve the generation effect of the large model. The specific instruction information is shown in the Table \ref{tab:findings}.

Then, the output is a new dataset in the target new scenario, i.e., the set of possible behaviors of the user in the new scenario, along with explanation generated by the corresponding changes

For example, based on a piece of data in spring [Monday, (18-21), Light, on, Monday, (18-21), Airconditioner, switch on, Monday, (18-21), Airconditioner, cooling, Monday, (21-24), Airconditioner, switch off, Monday, (21-24), Fan, switch on,...], the LLM  generates a new synthetic data piece for the target winter scene: [Monday, (15-18), Light, on, Monday, (18-21), Heater, switch on, Monday, (18-21), Heater, heating, Monday, (18-21), TV, switch on, Monday, (18-21), TV, setVolume,...]. 

In this example, the LLM recognizes the seasonal transition to winter and adapts the behavior sequence accordingly. It adjusts lighting times to reflect shorter daylight hours, replaces cooling devices with heating, and introduces more indoor entertainment activities. \schemename . Finally, extracts the new user behavior sequence dataset from the output and transform the text into item to facilitate the adaptive training of subsequent downstream task models.

\section{Conclusion}
The rise of smart home systems highlights the need for adaptable security solutions that can handle dynamic environments. Traditional methods, relying on static data, fail to address changing user behaviors and privacy concerns. This paper introduces \schemename, a novel framework that uses LLMs to generate synthetic datasets, tackling these issues. The Structure Pattern Perception Compression (SPPC) method, based on autoencoders, efficiently reduces token consumption while preserving data integrity. We also present a systematic approach for prompt construction to generate accurate IoT synthetic data. Our work advances secure, intelligent, and privacy-aware smart home systems.

\bibliographystyle{IEEEtran}
\bibliography{reference}

\vfill

\end{document}